\documentclass{article}

\usepackage{PRIMEarxiv}

\usepackage[utf8]{inputenc} % allow utf-8 input
\usepackage[T1]{fontenc}    % use 8-bit T1 fonts
\usepackage{hyperref}       % hyperlinks
\usepackage{url}            % simple URL typesetting
\usepackage{booktabs}       % professional-quality tables
\usepackage{amsfonts}       % blackboard math symbols
\usepackage{nicefrac}       % compact symbols for 1/2,  etc.
\usepackage{microtype}      % microtypography
\usepackage{lipsum}
\usepackage{fancyhdr}       % header
\usepackage{graphicx}       % graphics
\usepackage{subfigure}
\graphicspath{{media/}}     % organize your images and other figures under media/ folder
\usepackage{url}
\usepackage[ruled, vlined]{algorithm2e}
\include{pythonlisting}
\usepackage{indentfirst}
\usepackage{caption}  % The space between the table and its caption is very small
\usepackage{multirow}
\title{Paragraph2Graph: A GNN-based framework for layout paragraph analysis
}

\author{
  Shu Wei\\
  Datagrand Tech Inc. \\
  \texttt{weishucv@gmail.com} \\
  \And
   Nuo Xu\\
  Datagrand Tech Inc. \\
  \texttt{nxu8@outlook.com} \\
  \And
    Deng Huang\\
  Datagrand Tech Inc. \\
  \texttt{yangyu4608@gmail.com} \\
    \And
    Xiang Gao\\
  Datagrand Tech Inc. \\
  \texttt{gaoxiang@datagrand.com} \\
}

\begin{document}
\maketitle

\begin{abstract}
Document layout analysis has a wide range of requirements across various domains, languages, and business scenarios. However, most current state-of-the-art algorithms are language-dependent, with architectures that rely on transformer encoders or language-specific text encoders, such as BERT, for feature extraction. These approaches are limited in their ability to handle very long documents due to input sequence length constraints and are closely tied to language-specific tokenizers. Additionally, training a cross-language text encoder can be challenging due to the lack of labeled multilingual document datasets that consider privacy. Furthermore, some layout tasks require a clean separation between different layout components without overlap, which can be difficult for image segmentation-based algorithms to achieve. In this paper, we present Paragraph2Graph, a language-independent graph neural network (GNN)-based model that achieves competitive results on common document layout datasets while being adaptable to business scenarios with strict separation. With only 19.95 million parameters, our model is suitable for industrial applications, particularly in multi-language scenarios. We are releasing all of our code and pretrained models at \href{https://github.com/NormXU/Layout2Graph}{this repo}.

\end{abstract}

% keywords can be removed
\keywords{GNN \and Language-independent \and Document Layout \and Layout Paragraph \and Generalization}

\begin{figure}[!htbp]
    \centering
    \includegraphics[width=\linewidth]{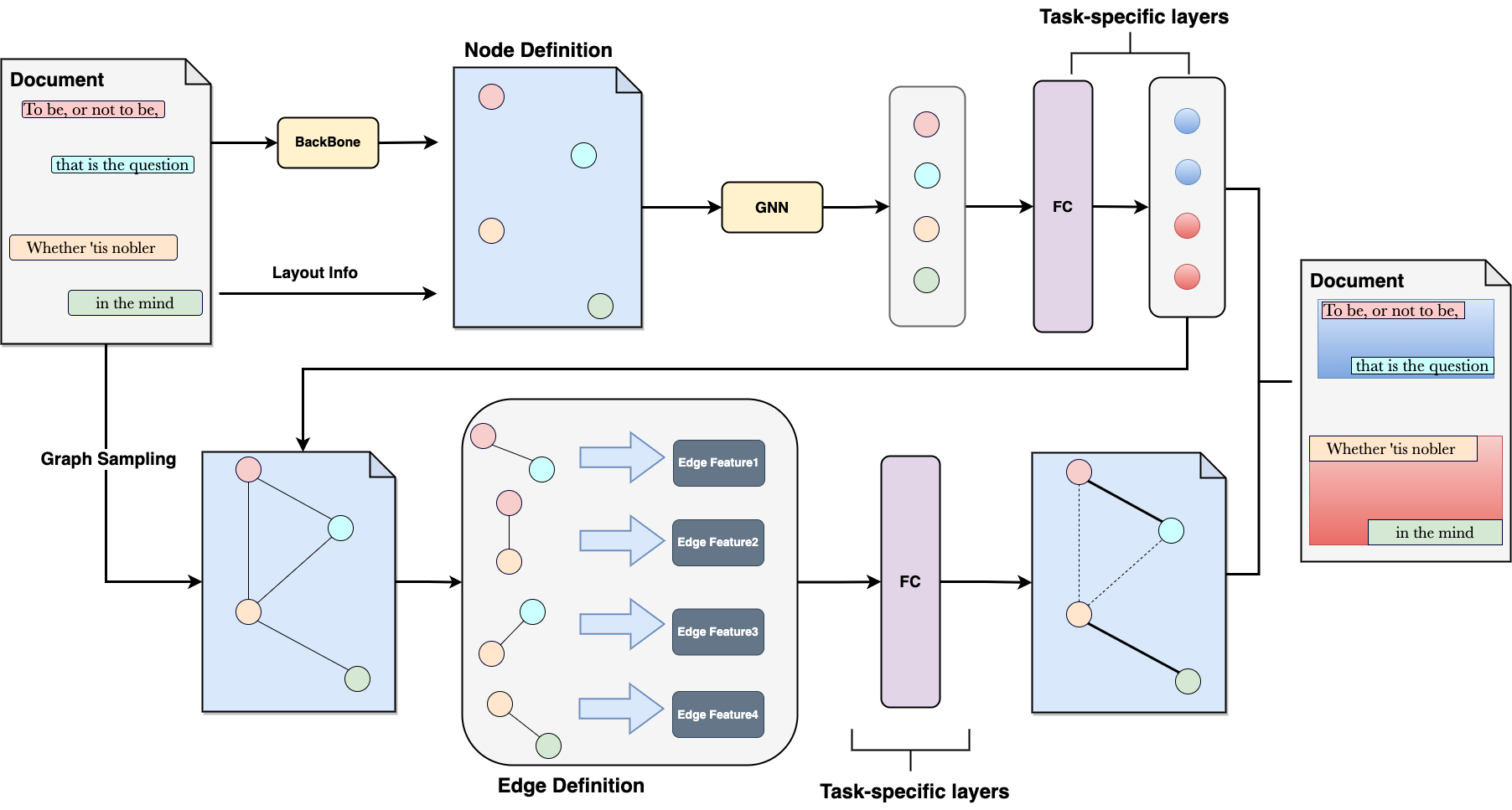}
    \caption{\textbf{The overall Paragraph2Graph architecture}. The whole pipeline consists of five parts: node definition, edge definition, GNN, graph sampling and task-specific layers;dotted lines represent invalid edge connections}
    \label{fig:overall}
\end{figure}

\section{Introduction}
Document layout analysis is an important task for very-rich document understanding. Given the availability to the text bounding boxes,  text info and document image,  most current works either integrate all modalities together with BERT-like encoders \cite{xu2020layoutlm}\cite{Xu2020LayoutLMv2MP}\cite{huang2022layoutlmv3}\cite{Hong2021BROSAL} or simply using visual information \cite{Ren2015FasterRCNN} \cite{he2017mask-rcnn} to model the task as an object detection problem. While effective,  industrial applications need to consider very-long multilingual paragraphs,  which a BERT-like encoder fails to hold due to the limitation of input sequence length and lack of multilingual document dataset. Moreover,  some scenarios expecting a clear separation between layout components make image segmentation-based algorithms hard to adapt due to vague boundaries. Although post-processing can handle the problems,  hand-craft rules make the pipeline complicated and hard to be maintained. In contrast,  graph neural networks (GNNs) can offer a promising alternative approach that does not rely on language models.

With this work, we propose Paragraph2Graph,  a language-independent GNN model to address these limitations. Fig.\ref{fig:overall} shows the overall architecture. We first encode image features with a pre-trained CNN backbone. Since each OCR box can be regarded as a spatially-separated node of a graph,  we therefore incorporate the 2d OCR text coordinates,  denoted as layout modality,  and image features together as node features. Then,  we build our neural network with DGCNN \cite{wang2019DGCNN} to dynamically refresh the graph based on updated node features and layout modality. As for edge features,  besides simply concatenating two node features,  relationship proposals \cite{zhang2017relationship} is also used for better capturing the relative spatial relationship. To improve the computation efficiency and balance between positive and negative training pairs,  we also propose a graph sampling method based on layout modality. A sparse graph can benefit forward and backward computations compared to fully-connected graphs. Finally,  two linear probes are trained to conduct node classification and edge classification respectively.

Our method does not require the use of a tokenizer or language model to extract text features as part of node embedding, making it language-independent and efficient in terms of parameters. In contrast to Transformer Encoder series or previous GNN works \cite{liu2019graph-conv}\cite{named}\cite{lee2021rope}\cite{luo-etal-2022-doc-gcn}\cite{gemelli2022doc2graph},  we have shown that Paragraph2Graph can easily generalize to multilingual documents without any modifications. We also conducted experiments showing that a model trained on Chinese documents performs similarly or even better on an English evaluation dataset than a model trained on English documents. This demonstrates the language-independence of our approach and indicates that the diversity of document layouts is the primary factor affecting performance. Additionally, our GNN models exhibit better generalization than object detection frameworks such as Faster-RCNN and Mask-RCNN.

Our contributions can be summarized as follows:
\begin{itemize}
 \item We propose a language-independent GNN framework which we call the Paragraph2Graph. The framework consists of node definition,  edge definition,  graph sampling,  GNN and task-specific layers. Each part of it can be easily edited. 
 
 \item We offer an empirical selection for each part of the Paragraph2Graph that achieves competitive results on several document layout analysis tasks.

\item We conduct extensive experiments and give our ablation analysis to justify the effective design.

\item The language-independent design allows us to make use of all public datasets to train a model regardless of language.

\end{itemize}

\begin{figure}[!htbp]
    \centering
    \includegraphics[width=\linewidth]{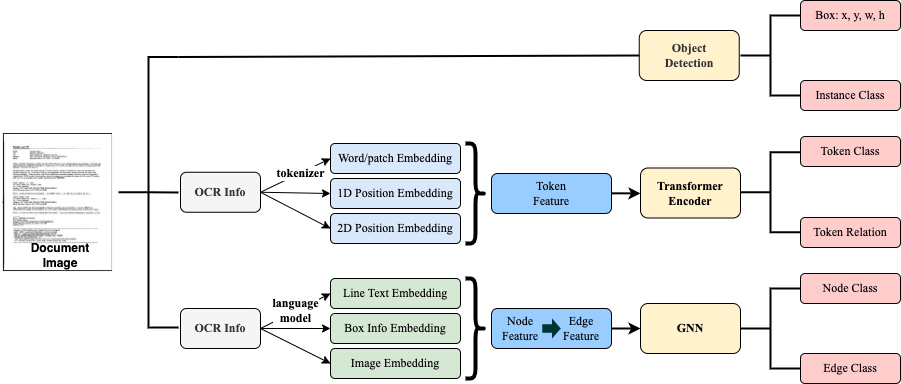}
    \caption{Three common practices for document layout analysis.}
    \label{fig:comm_practice}
\end{figure}

\section{Related Work}

Fig.\ref{fig:comm_practice} shows three common practices for  document layout analysis.

\subsection{Layout Tasks use Transformer Encoder}
A very-rich document has different modalities available across text info,  text position and image. To mimic how humans read,  LayoutLM \cite{xu2020layoutlm}, LayoutLMv2 \cite{Xu2020LayoutLMv2MP}, BROS \cite{Hong2021BROSAL} first integrate text features of each token with layout modality and corresponding image features. Afterward,  LayoutLMv3 \cite{huang2022layoutlmv3} extends the visual backbone to visual transformers. These frameworks define the document layout analysis as a relation extraction and follow the same design by constructing a fully-connected graph to calculate the relation score between all nodes. Then,  each relationship in the graph is inferred based on evaluating whether the score is over a threshold or not. All tokens inferred to be relational are grouped as one region. However,  these methods are highly language-dependent because of their language-specific tokenizers.  Due to the lack of labeled multilingual document datasets without privacy consideration, it is a challenging alternative plan to train a cross-language tokenizer. Besides, with the limitation of input sequence length,  Transformer Encoder-series methods cannot deal with very-long documents,  such as dense tables in financial reports. The high computational cost introduced by self-attention and fully-connection graphs reduce the availability to the industry as well.  

\subsection{Layout Tasks use Object Detection}
\begin{figure}[!htbp]
\centering  %图片全局居中
\subfigure[ground truth layout regions]{
\label{Fig.1}
\includegraphics[width=0.45\linewidth]{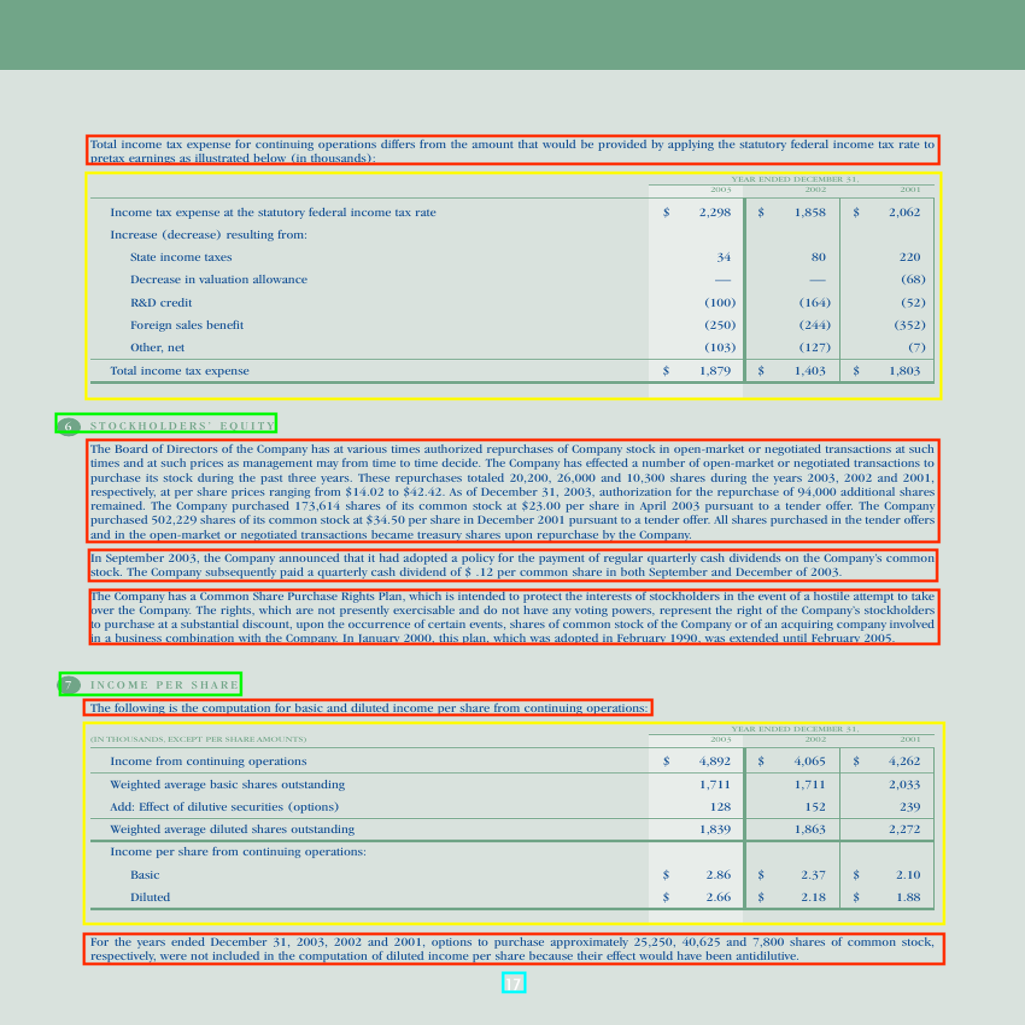}}
\subfigure[ predicted regions by an object detection method]{
\label{Fig.2}
\includegraphics[width=0.45\linewidth]{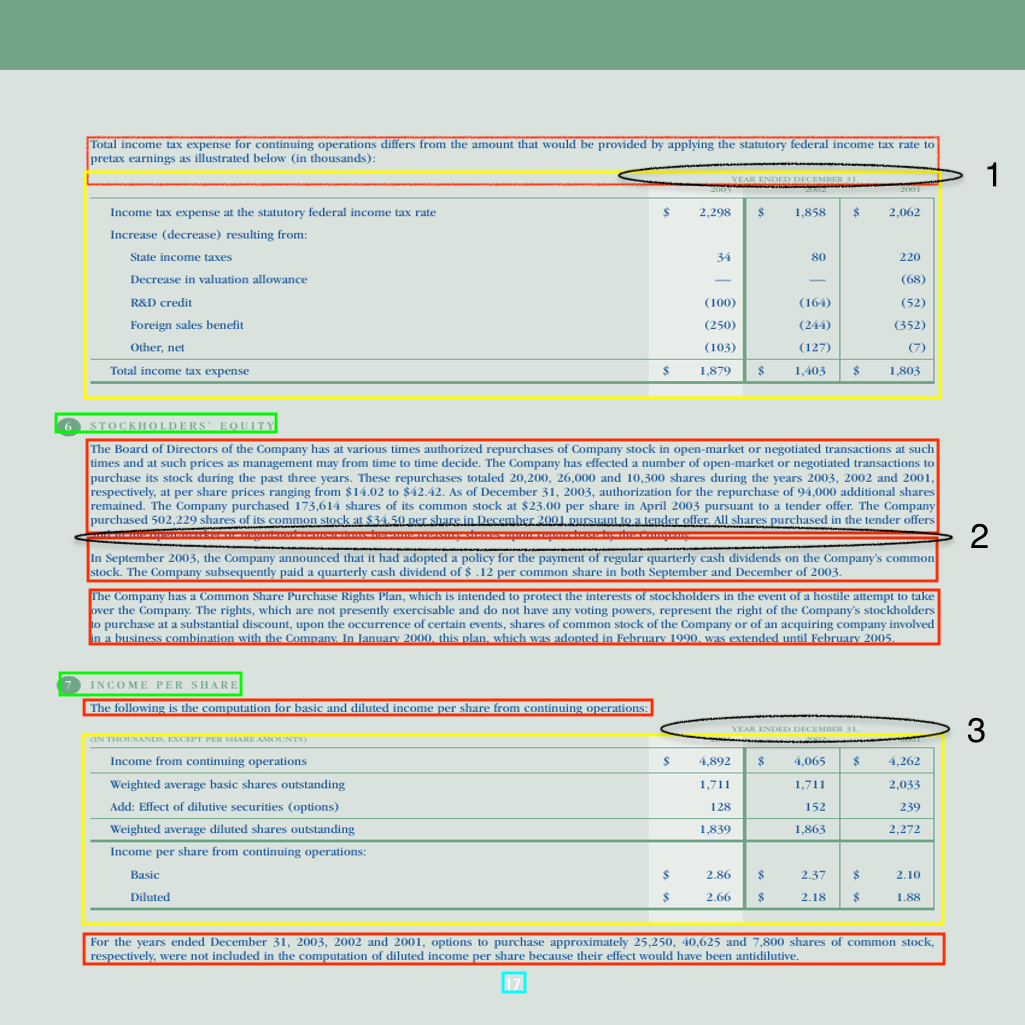}}
\caption{(a) ground truth layout region (b) tricky cases that an object detection method fails to handle with: \textbf{region 1}: shows one text box can locate across two layout regions;\textbf{region 2}: the text box exactly located at the box boundary;\textbf{region 3}: The text box is not located in any regions. green, red, yellow, blue rectangle means detected regions for the Title, Text, Table, and Page Foot.}
\label{fig:od-fail}
\end{figure}

Document layout analysis is about detecting the layout location of unstructured digital documents by returning bounding boxes and categories such as figures,  tables,  headers,  footers,  paragraphs,  etc. Such a task is initially defined as an object detection problem on which many algorithms \cite{Ren2015FasterRCNN}\cite{he2017mask-rcnn} \cite{girshick2014rich} related to object detection or segmentation have been successfully applied. However,  for all the object detection or segmentation models,  the predicted bounding boxes may overlap with each other due to the vague boundary between instances as shown in Fig.\ref{fig:od-fail}. The slight offset of the prediction boxes has little effect on the training loss, which in turn contributes limited to the model optimization to reach a high IoU. It's hard to assign a label to a text box that is either located at the edge of a predicted region or is shared by multiple predicted regions,  which makes the $AP^{IoU\geq0.9}$ less satisfying. It has to be mentioned that recent works \cite{huang2022layoutlmv3}\cite{Li2022DiTSP}\cite{Gu2021UniDocUP} replacing CNN backbone with vision transformer to achieve state-of-art results on public datasets,  but the uncompetitiveness of computation cost can't be ignored.

\subsection{Layout Task use GNN}
Graph Neural Networks (GNNs) have their special advantages in modeling spatial layout patterns of documents. Each text box of a document can be regarded as a spatially-separated node in a graph;text boxes grouped in the same layout region can be seen as being connected by edges. Since the document layout analysis can be implemented as node and edge classification from a sampled graph,  there exist no vague text boxes hard to be assigned to a certain group. We conclude a general pipeline covering all existing GNN-based layout analysis algorithms——node definition,  graph sampling,  edge definition,  GNNs and task-oriented processing. Existing works only explore part of the pipeline.

For node definition,  \cite{liu2019graph-conv} uses char embedding with BiLSTM to integrate text into node;\cite{Michael2020ICPR2C} adds regional image feature from FPN output;\cite{named} encodes box coordinate $xywh$ into node embedding. Afterward,  Post-OCR \cite{Wang2021PostOCRPR} adds $cos\alpha, sin\alpha, x cos\alpha, xsin\alpha, ycos\alpha, ysin\alpha$ and the width of the first word,  where $\alpha$ is the angle of text box;ROPE\cite{lee2021rope} explores the importance of the reading orders of given word-level node representations in a graph;Doc2Graph \cite{gemelli2022doc2graph} uses a pre-trained U-Net to get text image feature;Doc-GCN \cite{luo-etal-2022-doc-gcn} proposes a large collection for node definition,  which includes text embedding from BERT model,  image feature from pre-trained Faster-RCNN model,  the number of tokens,  the ratio of token number and box area,  and syntactic feature.

For edge definition, \cite{liu2019graph-conv} uses horizontal and vertical distance and the ratio of height between the two text boxes;ROPE\cite{lee2021rope} constructs with spatial embeddings from horizontal and vertical normalized relative distances between centers,  top left corners, and bottom right corners and relative height and width aspect ratios;Doc2Graph\cite{gemelli2022doc2graph} uses the output of the last GNN layer,  softmax of the output logits and polar coordinates for node embedding.

For the GNN module,  based on the vanilla GNNs,  many works have studied sophisticated designs to improve GNNs performances. Graph Convolutional Networks \cite{Kipf2016GCN} is a type of Graph Neural Network which applies convolution over graph structures. This design is widely used in \cite{luo-etal-2022-doc-gcn}\cite{Liu2022UnifiedLA}. GAT\cite{velivckovic2017gat} leverages the self-attention mechanism into GNNs to decouple node update coefficients from the structure of the graph. It has been used in \cite{Michael2020ICPR2C}\cite{named}\cite{Wang2021PostOCRPR}.

For graph sampling,  given that a document usually has a large number of text boxes that can be regarded as nodes,  it is essential to construct a graph with both high connectivity and sparsity compared to a fully-connected graph to allow necessary gradient propagation. \cite{kirkpatrick1985framework} first proposes $\beta$-skeleton graph;GraphSage \cite{hamilton2017inductive} uniformly samples a set of nodes from the neighborhoods and only aggregates feature information from sampled neighbors. \cite{Wang2021PostOCRPR}\cite{lee2021rope}\cite{Liu2022UnifiedLA} all follow $\beta$-skeleton to build their graph,  but the miss to cover tabular structures where text box density is relatively high. K-Nearest Neighbor is another good substitution, \cite{named} set $K=10$ and \cite{deep-relational} set $K=3$,  but it is still too tricky to tune satisfying parameters for different business scenarios.

\subsection{Other Tasks use GNN: table recognition, text line grouping}
All aforementioned algorithms can be generalized to table recognition tasks by simply modifying the task-oriented layers to represent whether two adjacent cells are in the same row or col. For the table recognition task, \cite{Li2020GFTEGF} uses KNN to construct a graph and represent text features by encoding character embedding with GRU;\cite{8978027-ReS2TIM} constructs a fully-connection graph and set weighted loss to balance between positive and negative samples.\cite{Chi2019GraphTSR} \cite{Qasim2019RethinkingTR} \cite{Raja2020TableStruct-net} \cite{9709898-TGRNet} share the identical GNN structure. Text line grouping task is  more easily adaptable to GNN with minor changes. \cite{deep-relational} predicts edge classified probability to judge if the pivot and its neighbors are in the same line;\cite{Riba2019Table-Invoice} introduces the residual connection mechanism for GNNs. In general,  a powerful GNN model can be used in many downstream document analysis tasks.

\section{Method}
 We follow our conclusion to establish a unified pipeline covering all main steps to build a GNN-based model for layout analysis: they include node definition,  graph sampling,  edge definition,  GNNs, and task-oriented processing.

\paragraph{Node definition}
Given a document image $D \in \mathbb{R}^{H\times W \times 3}$ with $N$ text boxes generated by any commercial or open source  Optical Character Recognition 
 (OCR) engine. we denote all text boxes as: 
 $\texttt{position\_info} = \{x_{min}^n,  y_{min}^n,  x_{max}^n,  y_{max}^n \mid n \in [0, N-1] \}$. The input image D is first resized into $D^\prime \in \mathbb{R}^{400\times 400 \times 3}$. $D^\prime$ is sent to a pre-trained ResNet visual backbone to get a series of output features from different scales. These features are integrated with $D^\prime$ into $F \in \mathbb{R}^{400\times 400 \times d}$ with an FPN structure. For $N$ text boxes,  we pick out their corresponding image features with ROIAlign and embed them as $I^{N \times k}$,  aka image embedding. $d$,  $k$ is the intermediate dimensions.  Given the normalized bounding box,  the layout information is represented as $\texttt{layout}_n = (x_{min}^n,  y_{min}^n,  x_{max}^n,  y_{max}^n,  x_{ctr}^n,  y_{ctr}^n,  w_n,  h_n)$. The layout information composes of bounding box coordinates,  center point coordinates,  bounding box width and height to construct a token-level 2D positional embedding,  denoted as layout embedding. We then fuse layout embedding and image embedding:
 
$$ \texttt{node\_embedding} := \texttt{MLP}(\texttt{Concate}(\texttt{image\_embedding} ,  \texttt{layout\_embedding}) $$

\paragraph{GNN module}
After gathering all the node features,  they are passed as input to the interaction model. We have tested two graph neural networks to use as the interaction part which are the modified versions of \cite{wang2019DGCNN} and \cite{Qasim2019LearningRO-gravnet} respectively. These modified networks are referred to as DGCNN* and GravNet* hereafter. We update the node features by aggregating weighted neighbor nodes with DGCNN/GravNet.
 
$$ \texttt{node\_embedding} = \max(\texttt{node\_embedding},  \texttt{DGCNN or GravNet}(\texttt{node\_embedding},  \texttt{position\_info})) $$

\begin{figure}[!htbp]
\centering  %图片全局居中
\subfigure[left]{
\label{Fig.1}
\includegraphics[width=0.45\linewidth]{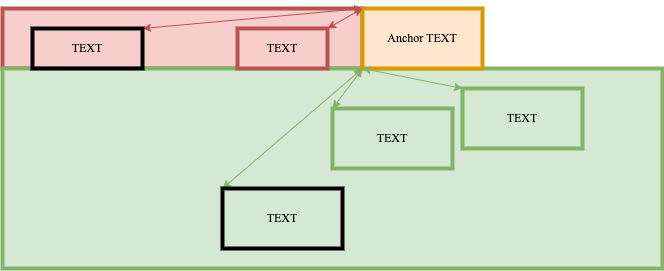}}
\subfigure[right]{
\label{Fig.2}
\includegraphics[width=0.45\linewidth]{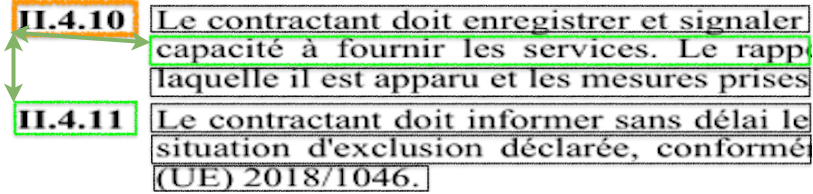}}
\caption{\textbf{Left}: illustration of our graph sampling strategy: for each node (shown in orange), we sample one edge horizontally (shown as red region) and two edges vertically (shown as green region);\textbf{Right}: We vertically sample top-2 nearest edges instead top-1 to ensure the connectivity with this common right-alignment paragraph structures }
\label{fig:custom}
\end{figure}

\paragraph{Graph sampling}
Text boxes classified into the same layout category can be regarded as having an edge between them. We refer to this task as node grouping: to infer whether there exists an edge between a node pair. To construct node pairs,  we therefore connect all potential edges between the nodes with a location-based node search algorithm. Instead of constructing a fully-connected graph,  our method can both save computation costs and improve training efficiency. Based on the common structure of a document,  each text node can have potential edge connections both vertically and horizontally. For each text box,  we pick up its top 1-2 location-nearest text boxes in four directions (top,  bottom,  left and right). Complex cases need additional processing as shown in Fig.\ref{fig:custom}-b. Our method can effectively sample a sparse graph without missing necessary node pairs. Comparison among KNN, $\beta$-skeleton, and our sampling methods can be found in Appendix Fig.\ref{fig:bad_case}.

\paragraph{Edge definition}
 We concatenate the node feature of each valid node pair as $F_{pair}$.  Inspired by ROPE \cite{lee2021rope},  we encode natural reading orders of words as $F_{rope}$ to help capture the better sequential presentation between nodes. A new reading order code is first assigned to neighbors with respect to each text box. Then,  a sinusoidal encoding matrix is applied to encode the reading order index. 

We also consider that the relationship between nodes is an important feature that has been ignored by \cite{zhang2017relationship}. Following the relationship proposal,  suppose two nodes have a potential relationship,  we denote one node as $S$,  a subject,  the other as $O$,  an object,  and the relationship as $R$. $\Delta(S,  O) = (t^{SO}_x,  t^{SO}_y,  t^{SO}_w,  t^{SO}_h,  t^{OS}_x,  t^{OS}_y)$,  where \\
\begin{equation}
t^{SO}_x = (x^S - x^O)/w^S,  \quad t^{SO}_y = (y^S - y^O)/h^S\\
\end{equation}
\begin{equation}
t^{SO}_w = \log(w^S/w^O),  \quad t^{SO}_h = \log(h^S/h^O)\\
\end{equation}
\begin{equation}
t^{OS}_x = (x^O - x^S)/w^O,  \quad t^{OS}_y = (y^O - y^S)/h^O
\end{equation}
$x^S$,  $y^S$,  $w^S$,  $h^S$ represent the center coordinates,  width, and  height of a subject box,  similarly denotations apply for $x^O$,  $y^O$,  $w^O$,  $h^O$. The coordinates of $R$ is the minimum bounding rectangle of $S$,  $O$,  which means
$$(x_{min}^R,  y_{min}^R,  x_{max}^R,  y_{max}^R) = \min(x_{min}^O,  x_{min}^S),  \min(y_{min}^O,  y_{min}^S),  max(x_{max}^O,  x_{max}^S),  \max(y_{max}^O,  y_{max}^S)
$$
The relationship feature $F_{rel}$ is defined as an concatenation of $\Delta(S,  O)$,  $\Delta(S,  R)$ and $\Delta(O,  R)$.
Finally,  the edge feature is formally represented as:
$$ F_{edge}= \texttt{Concate}(F_{pair},  F_{rope},  F_{rel}) $$

\paragraph{Task-oriented processing}
For node classification,  we apply a fully-connected layer to fuse features and a linear layer $W^{h \times c}$ to classify each node. $h$ is the hidden dimension and $c$ is the number of categories. For node grouping,  we follow the same idea as node classification: a fully-connected layer to fuse features and a linear layer $W^{h \times 2}$ to infer whether there exists an edge between a node pair or not. All connected nodes are regarded as layout instance. The minimum bounding boxes of connected nodes are the final layout bounding boxes. The category mode of connected nodes is the category of the layout instance.

\section{Experiments}
Previous works propose various definitions on edge and nodes,  but they either get non-competitive results or introduce expensive computation costs. We therefore study the combinations of these designs and compare them with object detection and the Transformer Encoder model on several document layout analysis tasks. Our experiments demonstrate the effectiveness and competitiveness of our method.

We train our model with 1 GeForce 3090 GPUs from scratch, We use an Adam optimizer with 0.937 momentum and 0.005 weight decay. The learning rate is set to 0.0001.

\subsection{Results on Public Datasets}
\paragraph{FUNSD} The FUNSD \cite{jaume2019funsd} provides 199 annotated forms with 9, 707 entities and 31, 485 word-level annotations for four entity types: header,  question,  answer,  and other. It includes noisy scanned documents in English from various fields, such as research, marketing, and advertising. This dataset is commonly used in GNN-related papers, and we used it for our experiments for easy comparison. FUNSD contains two level labels: word and entity. For word-level labels, we predict the category of each word and determine whether two words belong to the same entity. For entity-level labels, we adds two classification heads: one for entity labeling and the other for entity linking, which predicts whether two entities are matched.
 
We report our best hyperparameter configuration as shown in Tab.\ref{table:ablation} ours-Large in the ablation experiment of section 4.3. We train the models with a batch size of 2 for 60 epochs and a warm-up period of 10 epochs. The training and validation set are split as provided, with 149 for training and 50 for evaluation.

To evaluate the performance of our method, we use multi-class F1-scores to for node classification and binary edge classification F1-scores for grouping or linking, along with corresponding precision and recall values. Our method significantly outperformed previous works. Despite not using a language model, our model had a significantly smaller number of parameters as shown in Tab.\ref{table:FUNSD-word}.

In the entity task,our model shown in Tab.\ref{table:FUNSD-entity}, achieved a F1 score of 0.80575 for entity-labeling and 0.77031 for entity-linking, using 32.98 million parameters. We achieved state-of-the-art results in entity linking, outperforming other GNN models and the Transformer Encoder series. On the entity labeling task, our F1 score was lower than some Transformer Encoder modeles such as LayoutLMv2, LayoutLMv3, and BROS. This may be due to those methods having more parameters and being pre-trained on several text-image alignment tasks, giving them strong semantic and visual understanding abilities. Despite being trained from scratch with only 149 samples, our model still outperformed BERT, Roberta, and LayoutLM, and achieves significant improvements over most previous GNN works. However, doc2graph performed $1.6\%$ better than our model on the entity labeling task, but it still suffers from the problems associated with language-based GNNs due to its use of a language model. Compared to other state-of-the-art models, our GNN model performed competitively. 

\begin{table}[!htbp]
\centering
\begin{tabular}{|c|c|c|c|c|}
\hline
& F1    & word-labeling    & word-grouping    & Params       \\ \hline
GNN with Language model & FUNSD\cite{jaume2019funsd} & -                & 0.41        & 340M         \\ 
GNN with Language model & Named\cite{named} & -                & 0.65        & 201M         \\ 
GNN with Language model & ROPE\cite{lee2021rope}  & 0.5722           & 0.8933           & -            \\ \hline
GNN & ours-Large  & \textbf{0.68933} & \textbf{0.91483} & \textbf{32.98M} \\ \hline
\end{tabular}

\caption{Performance for word level on FUNSD.}
\label{table:FUNSD-word}
\end{table}

\begin{table}[!htbp]
\centering
\begin{tabular}{|c|c|c|c|c|c|}
\hline
& F1           & entity-labeling  & entity-linking   & Params       \\\hline
GNN with Language model & FUNSD\cite{jaume2019funsd}        & 0.57             & 0.04             & 340M         \\
GNN with Language model & Named\cite{named}       & 0.64             & 0.39             & 201M         \\
GNN & FUDGE\cite{Davis2021VisualFUDGE}        & 0.6652           & 0.5662           & 17M         \\
GNN & Word-FUDGE\cite{Davis2021VisualFUDGE}   & 0.7221           & 0.6258                & 17M         \\
GNN with Language model & Doc2Graph\cite{gemelli2022doc2graph}    & 0.8225           & 0.5336           & 6.2M+ \\
Transformer Encoder & BERT-base\cite{Devlin2019BERTPO}    & 0.6026           & 0.2765           & 110M         \\
Transformer Encoder &BERT-L\cite{Devlin2019BERTPO}       & 0.6563           & 0.2911           & 340M         \\
Transformer Encoder &RoBERTa-base\cite{Liu2019RoBERTaAR} & 0.6648           &                  & 125M         \\
Transformer Encoder &RoBERTa-L\cite{Liu2019RoBERTaAR}    & 0.7072           &                  & 355M         \\
Transformer Encoder &LayoutLM\cite{xu2020layoutlm}     & 0.7927           & 0.4586           & 113M         \\
Transformer Encoder &LayoutLM-L\cite{xu2020layoutlm}   & 0.7789           & 0.4283           & 343M         \\
Transformer Encoder &LayoutLMv2\cite{Xu2020LayoutLMv2MP}   & 0.8276           & 0.4291           & 200M         \\
Transformer Encoder &LayoutLMv2-L\cite{Xu2020LayoutLMv2MP} & 0.8420           & 0.7057           & 426M         \\
Transformer Encoder &BROS\cite{Hong2021BROSAL}         & 0.8305           & 0.7146           & 138M         \\
Transformer Encoder &BROS-L\cite{Hong2021BROSAL}       & 0.8452           & 0.7701           & 340M         \\
Transformer Encoder &LayoutLMv3\cite{huang2022layoutlmv3}   & 0.9029           &                  & 133M         \\
Transformer Encoder &LayoutLMv3-L\cite{huang2022layoutlmv3} & \textbf{0.9208}  &                  & 368M         \\\hline
GNN & ours-Large         & 0.80575 & \textbf{0.77031} & \textbf{32.98M} \\\hline       
\end{tabular}
\caption{Performance  for entity level on FUNSD. Doc2Graph params counts a spaCy model and a pre-trained U-Net on FUNSD besides its own weights.}
\label{table:FUNSD-entity}
\end{table}

\paragraph{PublayNet}
PublayNet\cite{Zhong2019PubLayNetLD} contains research paper images annotated with bounding boxes and polygonal segmentation across five document layout categories: Text,  Title,  List,  Figure,  and Table. The official splits contain 335, 703 training images,  11, 245 validation images,  and 11, 405 test images. We train our model on the training split and evaluate our model on the validation split following standard practice. We train our models  with the batch size of 4  for 5 epochs and  warm-up 1 epoch. Because of training resource limitations, we only report our suboptimal configuration shown in Tab.\ref{table:ablation} ours-Small. 

\begin{figure}[!htbp]
\centering  %图片全局居中
\subfigure[finaly layout analysis output]{
\label{Fig.1}
\begin{minipage}[c]{\linewidth}
\includegraphics[width=0.3\linewidth]{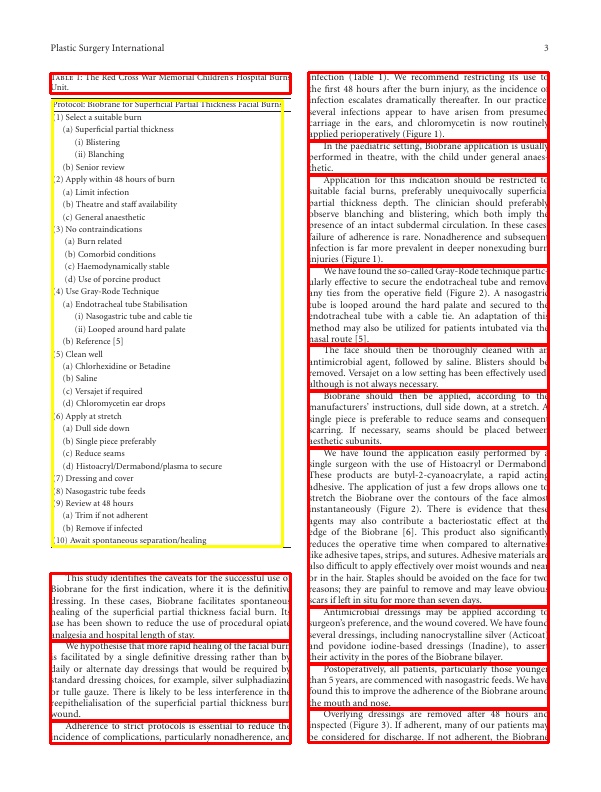}
\includegraphics[width=0.3\linewidth]{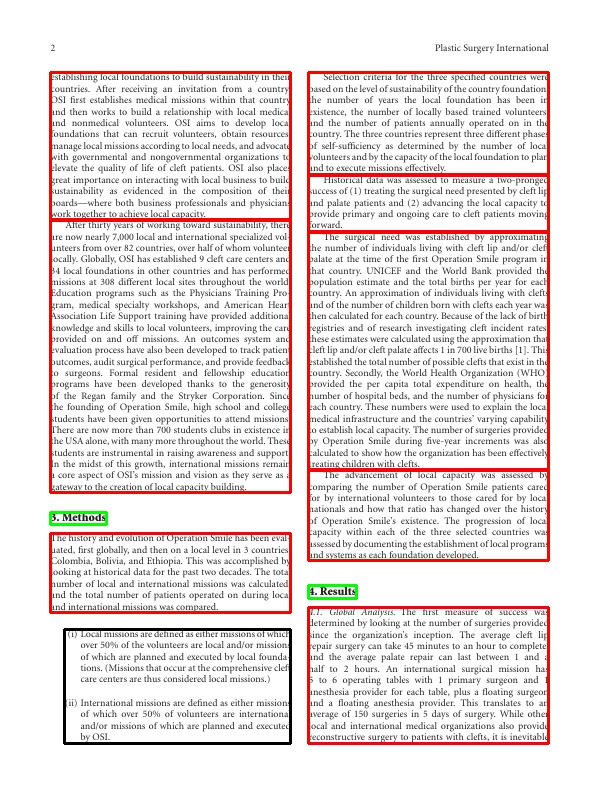}
\includegraphics[width=0.3\linewidth]{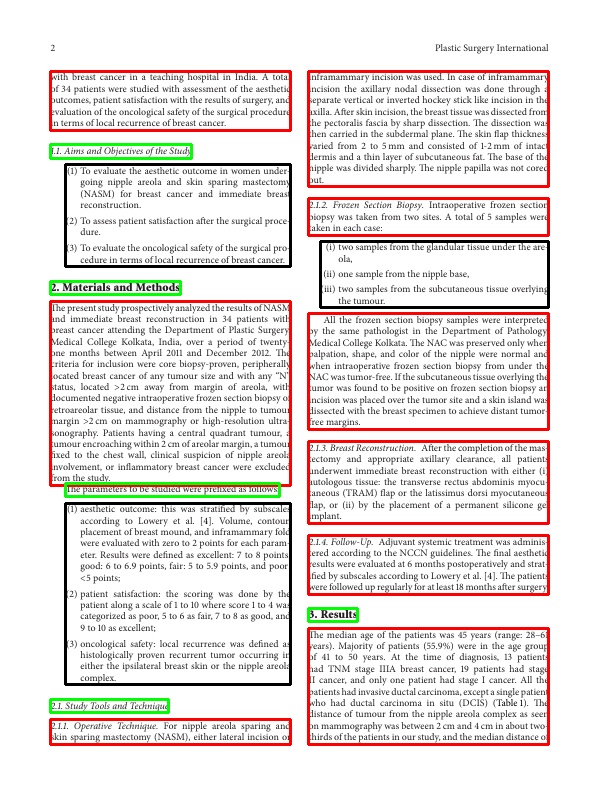}
\end{minipage}
}
\subfigure[connect all paired edges]{
\label{Fig.2}
\begin{minipage}[c]{\linewidth}
\includegraphics[width=0.3\linewidth]{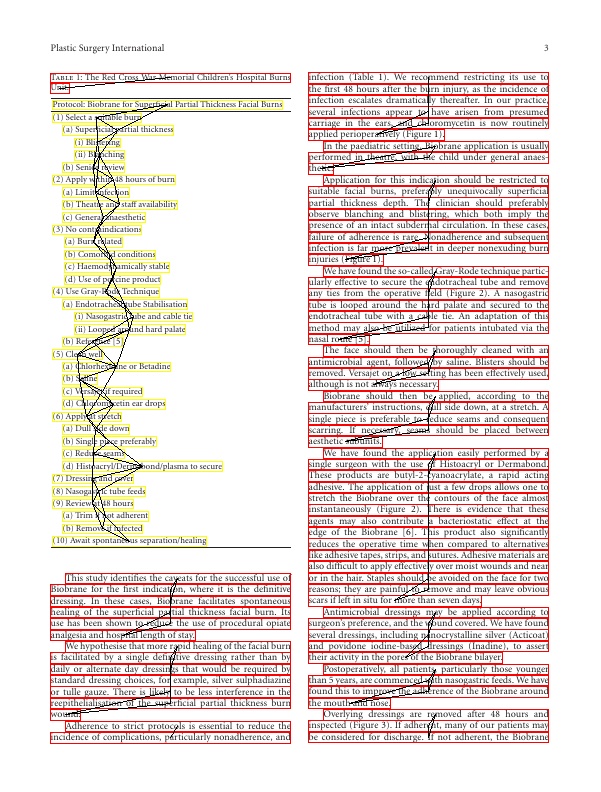}
\includegraphics[width=0.3\linewidth]{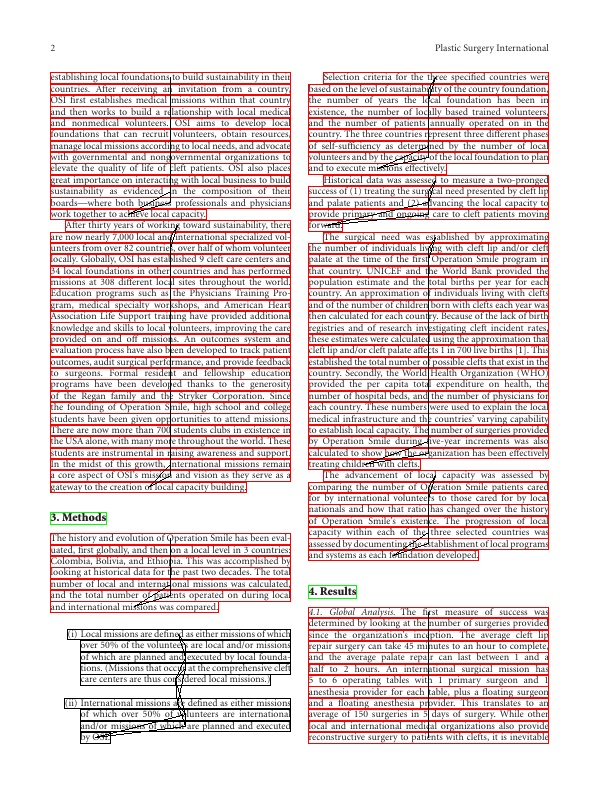}
\includegraphics[width=0.3\linewidth]{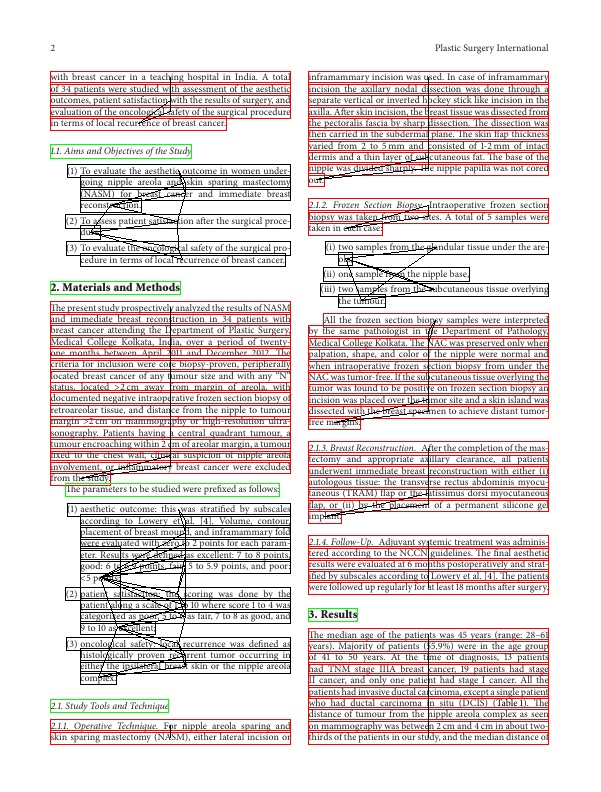}
\end{minipage}

}
\subfigure[connect all unpaired edges]{
\label{Fig.3}
\begin{minipage}[c]{\linewidth}
\includegraphics[width=0.3\linewidth]{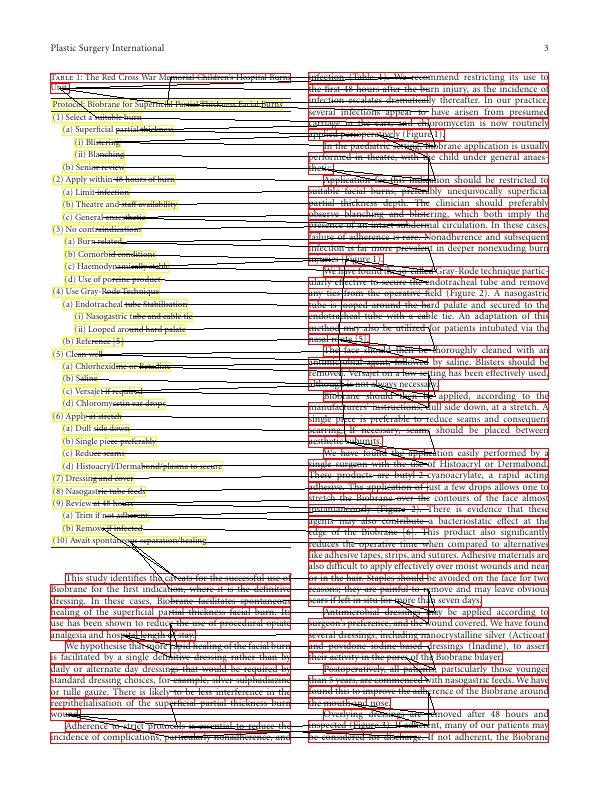}
\includegraphics[width=0.3\linewidth]{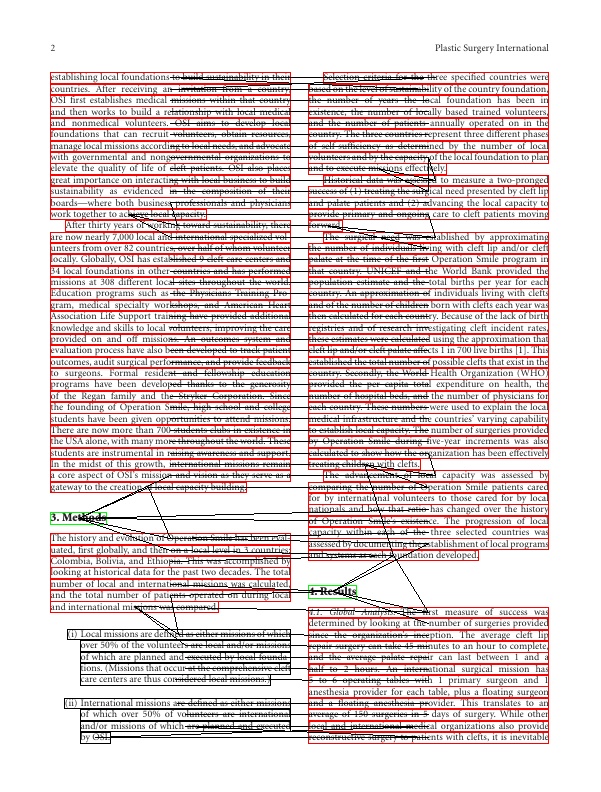}
\includegraphics[width=0.3\linewidth]{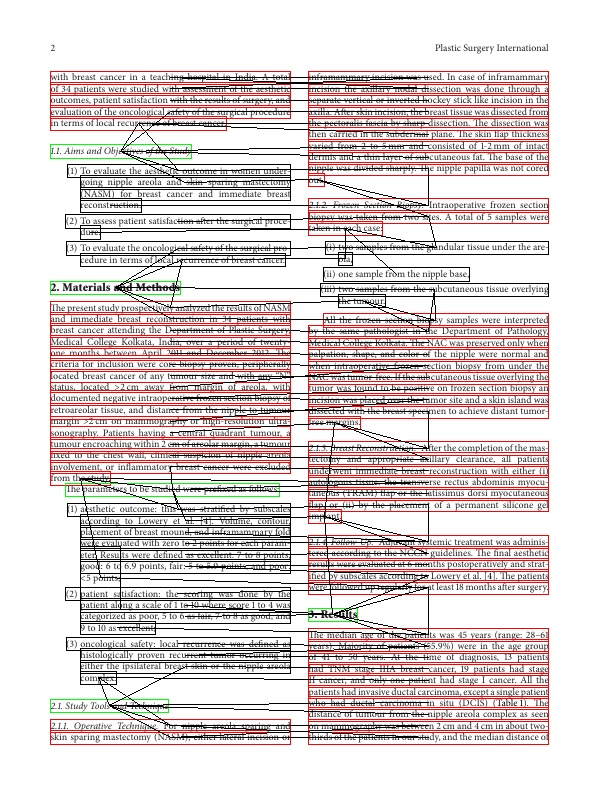}
\end{minipage}
}
\caption{Result on Publaynet: a) rectangle color red, green, yellow, black means Text, Title, Table, and List;black line means edge between two nodes;b) edge predicted to be connected;c) edge predicted to be unconnected.}
\label{fig:result}
\end{figure}

We measure the performance using the mean average precision (mAP) @ intersection over union (IOU) [0.50:0.95] of bounding boxes and report results in Tab.\ref{table:Publaynet} with only two categories : Text and Title. Tab.\ref{table:Publaynet-all} shows all categories. Our proposed GNN-based model outperforms several state-of-the-art object detection models. Specifically, our model achieves a mAP of 0.954 and 0.913 for Text and Title detection, with a model size of 77M. Our model shows better performance than Faster-RCNN,Cascade-RCNN and Mask-RCNN regardless of whether they have been pretrained , which sizes ranging from 168M to 538M. Compared to Faster-RCNN-Q and Post-OCR models  with mAPs of 0.914 and 0.892 respectively, our GNN-based model achieves higher accuracy in Text category. Fig.\ref{fig:result} shows some cases of our algorithm on this dataset.

\begin{table}[!htbp]
\centering
\begin{tabular}{|c|c|cc|c|}
\hline
&mAP                           & Text           & Title                 & Size \\\hline
OD & Faster-RCNN\cite{Zhong2019PubLayNetLD}                   & 0.910          & 0.826            & -      \\
OD & Mask-RCNN\cite{Zhong2019PubLayNetLD}                     & 0.916          & 0.84              & 168M   \\
OD-pretrained & Faster-RCNN{[}UDoc{]}\cite{Gu2021UniDocUP}         & 0.939          & 0.885                & -      \\
OD-pretrained & Mask-RCNN{[}DiT-base{]}\cite{Li2022DiTSP}       & 0.934          & 0.871           & 432M   \\
OD-pretrained & Cascade-RCNN{[}DiT-base{]}\cite{Li2022DiTSP}    & 0.944          & 0.889            & 538M   \\
OD-pretrained & Cascade-RCNN{[}layoutlm-v3{]}\cite{huang2022layoutlmv3} & 0.945          & 0.906            & 538M   \\
OD & Faster-RCNN-Q\cite{Wang2021PostOCRPR}                 & 0.914          &                & -      \\
GNN & Post-OCR\cite{Wang2021PostOCRPR}                      & 0.892          &         & -      \\ \hline
GNN & ours-Large                          & \textbf{0.954}          & \textbf{0.913}        & \textbf{77M}   \\\hline
\end{tabular}
\caption{Performance on Publaynet on paragraph categories.OD-pretrained means object model use pretrained CNN.}
\label{table:Publaynet}
\end{table}

\paragraph{Doclaynet}
Doclaynet\cite{Pfitzmann2022DocLayNetAL} is a recently released document layout  dataset annotated in COCO format. It contains 80863 manually annotated pages from diverse data sources to represent a wide variability in layouts with 11 distinct classes: Caption,  Footnote,  Formula,  List-item,  Page-footer,  Page-header,  Picture, Section-header,  Table,  Text,  and Title. Compared with publaynet,  this dataset covers more complex and diverse document types,  including Financial Reports,  Manuals,  Scientific Articles,  Laws \& Regulations,  Patents and Government Tenders.

We use the same training parameters as in PubLayNet and evaluate the quality of their predictions using mean average precision (mAP) with 10 overlaps that range from 0.5 to 0.95 in steps of 0.05 (mAP@0.5-0.95). These scores are computed by leveraging the evaluation code provided by the COCO API. Similarly,we only compare categories belonging to the Paragraph type without Table and Picture.The results shown in Tab.\ref{table:Doclaynet} draw a similar conclusion as we do in Publaynet: ours achieves better results with only 1/7 parameters with respect to object detection models in total with a mAP of 0.771. Specifically, our model performs exceptionally well in Page-Header, Caption, Section-Header, Title, and Text detection with mAPs of 0.796, 0.809, 0.824, 0.643, and 0.827, respectively. Comparatively, YOLO-v5x6 achieves best result in Footnote, Title and Text.Tab.\ref{table:Doclaynet-all} shows all categories.

\begin{table}[!htbp]
\centering
\setlength{\parindent}{-4em}
\scalebox{0.7}{
\begin{tabular}{|c|c|c|c|c|c|}
\hline
mAP & Mask-RCNN-res50\cite{Pfitzmann2022DocLayNetAL}  & Mask-RCNN-resnext101\cite{Pfitzmann2022DocLayNetAL} & Faster-RCNN-resnext101\cite{Pfitzmann2022DocLayNetAL} & YOLO-v5x6\cite{Pfitzmann2022DocLayNetAL}   & ours-Small                       \\ \hline
Page-Header     & 0.719                                                           & 0.7                                                                  & 0.720                                                                  & 0.679                                                     & \textbf{0.796}  \\ 
Caption         & 0.684                                                           & 0.715                                                                & 0.701                                                                  & 0.777                                                     & \textbf{0.809}  \\ 
Formula         & 0.601                                                           & 0.634                                                                & 0.635                                                                  & 0.662                                                     & \textbf{0.726}  \\ 
Page-Footer     & 0.616                                                           & 0.593                                                                & 0.589                                                                  & 0.611                                                     & \textbf{0.920}  \\ 
Section-Header  & 0.676                                                           & 0.693                                                                & 0.684                                                                  & 0.746                                                     & \textbf{0.824}  \\ 
Footnote        & 0.709                                                           & 0.718                                                                & 0.737                                                                  & \textbf{0.772}                           & 0.625                            \\ 
Title           & 0.767                                                           & 0.804                                                                & 0.799                                                                  & \textbf{0.827}                           & 0.643                            \\ 
Text            & 0.846                                                           & 0.858                                                                & 0.854                                                                  & \textbf{0.881}                           & 0.827        
                          \\ \hline
Total-paragraph & 0.702                                                           & 0.7143                                                               & 0.7148                                                                 & 0.744                                                     & \textbf{0.771}  \\ \hline
Params          & -                                                               & -                                                                    & 60M                                                                    & 140.7M                                                    & \textbf{19.95M} \\ \hline
\end{tabular}
}
\caption{performance on Doclaynet on paragraph categories.}
\label{table:Doclaynet}
\end{table}

\subsection{Discussion on Generalization of Language}
To demonstrate the language-independence of our model,  we first train our model on datasets in English and evaluate them on a dataset in Chinese. Among them, Publaynet is a pure English dataset, Doclaynet is mostly in English, and DGDoc is a pure Chinese dataset containing 12, 000 images from real business scenarios. We use the same F1 indicator as the experiment on FUNSD. As shown in Tab.\ref{table:language-independent}, our models trained on Doclaynet data and models trained on Chinese-data behave similarly on Publaynet,  and even the latter one gets a higher F1 on two tasks. The model trained on Publaynet data and the model trained on Chinese-data perform similarly on Doclaynet. This experiment proves that our model has the language-independent ability,  which allows us to focus more on training data with diversity and complex layout structures instead of languages. The most important advantage of this conclusion is that for non-English application scenarios,  we do not need to collect and annotate a large number of documents,  which is very time-consuming and expensive. Instead, we can directly collect a variety of public datasets regardless of languages to train the model.

\begin{table}[!htbp]
\centering
\begin{tabular}{|c|cc|cc|cc|}
\hline
\multirow{2}{*}{F1} 
& \multicolumn{2}{c|}{val-doclaynet}  & \multicolumn{2}{c|}{val-publaynet}  & \multicolumn{2}{c|}{val-chineseData}   \\
& node & edge & node & edge & node & edge \\\hline
train-doclaynet     & 0.94670                  & 0.97267                  & 0.95774                  & \textbf{0.97171}                  & 0.85892                  & 0.94851                  \\
train-publaynet     & 0.72742                  & 0.86069                  & 0.98729                  & 0.99302                  & 0.76158                  & 0.8839                   \\
train-chineseData   & \textbf{0.88256}                  & \textbf{0.92285}                  & \textbf{0.96289}                  & 0.96847                  & 0.97295                  & 0.98872       \\\hline          
\end{tabular}
\caption{Comparison results for discussion on generalization of language.}
\label{table:language-independent}
\end{table}

\subsection{Discussion on Generalization of Data Complexity}
In order to compare the generalization of the model, we use two datasets:Doclaynet is a more diverse dataset than Publaynet in layout. As shown in Tab.\ref{table:data-complexity}, if we trained on Publaynet and predicted on Doclaynet, both models Mask-RCNN or our GNN-based model drop badly in mAP. But if we use the model trained on Doclaynet to predict on Publaynet, our model only slightly decreases, while Mask-RCNN drops significantly. It shows that as long as our model has been trained on complex and diverse layouts, it can also migrate to simple layouts well.

\begin{table}[!htbp]
\centering
\begin{tabular}{|cc|cc|cc|}
\hline
\multicolumn{2}{|c|}{\multirow{2}{*}{mAP}}                              & \multicolumn{2}{c|}{val-publaynet} & \multicolumn{2}{c|}{val-doclaynet} \\
\multicolumn{2}{|c|}{}                                                  & maskrcnn-res50     & ours-Small    & maskrcnn-res50   & ours-Small      \\ \hline
\multicolumn{1}{|c|}{\multirow{2}{*}{train-publaynet}} & Section-Header & 0.87               & 0.913         & 0.32             & \textbf{0.484}  \\
\multicolumn{1}{|c|}{}                                 & Text           & 0.96               & 0.954         & 0.42    & 0.338           \\ \hline
\multicolumn{1}{|c|}{\multirow{2}{*}{train-doclaynet}} & Section-Header & 0.53               & \textbf{0.811}         & 0.68    & 0.796           \\
\multicolumn{1}{|c|}{}                                 & Text           & 0.77               & 0.769         & 0.84             & 0.827           \\ \hline
\end{tabular}
\caption{Comparison results for discussion on generalization of data complexity.}
\label{table:data-complexity}
\end{table}

\subsection{Ablation Experiments}
\subsubsection{Node-definition}
When considering the node definition, we tested whether the box information is 4 values or 8, the backbone tried res-18 and res-50, GNN tried DGCNN and GravNet, whether the GNN input and output are residual connect.The impact of each factor was compared on four tasks for more comprehensive assessment in Tab.\ref{table:ablation-node}.The addition of residual connections (using res-18 or res-50) generally leads to better performance than non-residual networks.The use of more big backbone (res-50 vs res-18) can improve the results for all tasks.When comparing different GNN models, DGCNN tend to perform better than Gravnet.However the importance of box information is vague.
\begin{table}[!htbp]
\centering
\setlength{\parindent}{-4em}
{
\begin{tabular}{|cccc|cc|cc|}
\hline
box-infor & backbone & GNN & GNN-res    & WL               & WG               & EL               & EG               \\\hline
4         & res-18   & D   & -         & 0.65189          & 0.86821          & 0.76458          & 0.67797          \\
8         & res-18   & D   & -          & 0.63914$\downarrow$          & 0.86248$\downarrow$          & 0.77573          & 0.69695          \\
8         & res-18   & D   & +   & 0.65579          & 0.86399$\downarrow$          & 0.77144          & 0.70509          \\
8         & res-50   & D   & -     & 0.66774          & 0.87487          & 0.77873          & 0.71613          \\
8         & res-18   & G   & -    & 0.58861$\downarrow$          & 0.84461$\downarrow$          & 0.71397$\downarrow$          & 0.60032$\downarrow$          \\\hline
\end{tabular}
}

\caption{Ablation experiments of node definition on FUNSD: WL, WG, EL, EG means word-labeling, word-grouping, entity-labeling, entity-linking;D and G means DGCNN and Gravnet;box-info 4 means $x_{min}, y_{min}, w, h$, 8 means $x_{min}, y_{min}, x_{max}, y_{max}, x_{center}, y_{center}, w, h$.}
\label{table:ablation-node}
\end{table}

\subsubsection{Edge-definition}
As for edge definition , we tested four factors:whether to use relationship,ROPE, polar and node class result as shown in Tab.\ref{table:ablation-edge}. Four tasks is also WL (word-labeling), WG (word-grouping), EL (entity-labeling), and EG (entity-linking). The results indicate that adding relation and node-class edges improves the performance of all metrics except EG. Adding the ROPE edge improves WL and WG but not EL and EG. Adding the polar edge does not significantly affect the results.
\begin{table}[!htbp]
\centering
\setlength{\parindent}{-4em}
{
\begin{tabular}{|cccc|cc|cc|}
\hline
relation\cite{zhang2017relationship} & ROPE\cite{lee2021rope} & polar\cite{gemelli2022doc2graph} & node-class\cite{gemelli2022doc2graph}  & WL               & WG               & EL               & EG               \\\hline
   &      &       &        & 0.63478          & 0.86348          & 0.77873          & 0.70071          \\
    &      &       & +      & 0.63914          & 0.86248$\downarrow$          & 0.77573$\downarrow$          & 0.69695$\downarrow$          \\
+        &      &       & +        & 0.63972          & 0.90039          & 0.77916          & 0.74759          \\
+        & +    & +     & +        & 0.63374$\downarrow$          & 0.89869$\downarrow$          & 0.77358$\downarrow$          & 0.74925          \\\hline
\end{tabular}
}

\caption{Ablation experiments  of edge definition on FUNSD: WL, WG, EL, EG means word-labeling, word-grouping, entity-labeling, entity-linking.}
\label{table:ablation-edge}
\end{table}

\subsubsection{Input Process \& Loss}
Tab.\ref{table:ablation-other} shows the results of various ablation experiments conducted on the FUNSD dataset, using different input processes and loss functions. The first row represents the baseline model, which uses a 400*400 image size and cross-entropy (CE) loss. The effect of image padding is investigated by adding a padding of unknown size to the input images. The results indicate that this modification does not significantly affect the word-labeling metric but has a negative impact on word-grouping, entity-labeling, and entity-linking. Increasing the image size to 800*608 pixels leads to some improvement in the word-grouping and entity-labeling metrics, but the word-labeling and entity-linking scores remain relatively low.Adding a contrastive loss term in addition to the cross-entropy loss leads to some improvement in all four tasks.Overall, Padding may not be helpful, larger image sizes may not lead to significant improvements, and the addition of a contrastive loss may be beneficial.
\begin{table}[!htbp]
\centering
\setlength{\parindent}{-4em}
{
\begin{tabular}{|ccc|cc|cc|}
\hline
image-size & image-pad & loss   & WL               & WG               & EL               & EG               \\\hline
 400*400    &           & CE     & 0.61066          & 0.86352          & 0.78216          & 0.72722          \\
 400*400    & +         & CE     & 0.63914          & 0.86248$\downarrow$          & 0.77573$\downarrow$          & 0.69695$\downarrow$          \\
800*608    & +         & CE     & 0.68795          & 0.87782          & 0.81947 & 0.70487$\downarrow$          \\
 400*400    & +         & CE+Con & 0.64178          & 0.87105          & 0.77658$\downarrow$          & 0.71704$\downarrow$    \\\hline
\end{tabular}
}

\caption{Ablation experiments of input process and loss on FUNSD: WL, WG, EL, EG means word-labeling, word-grouping, entity-labeling, entity-linking;CE and Con means CE loss and Contrastive loss.}
\label{table:ablation-other}
\end{table}

\subsubsection{Final Best Model}

\begin{table}[!htbp]
\centering
\setlength{\parindent}{-4em}
{
\begin{tabular}{|c|cc|cc|cc|}
\hline
Name   & backbone & image-size   & WL               & WG               & EL               & EG               \\\hline
ours-Small  & res-18         & 400*400      & 0.66579          & 0.90148          & 0.78216          & 0.75522          \\
ours-Large   & res-50         & 800*608      & \textbf{0.68933} & \textbf{0.91483} & \textbf{0.82504}          & \textbf{0.77031} \\\hline

\end{tabular}
}

\caption{Best model.}
\label{table:ablation-final}
\end{table}

We  study the combinations of a variety of designs for each component in the GNN model and justify their effectiveness on our document layout analysis tasks. 

As seen in Tab.\ref{table:ablation-final},  scale-up image size, relationship proposal, and larger CNN backbone are useful to improve accuracy,  while contrast loss, and Gravnet are not. The importance of other factors is ambiguous to tell. DGCNN is a better choice than Gravnet. Based on the detailed ablation experiment,  we get the best design combination is ours-Large. Since we enlarge the input image size,  more memory is cost. Therefore, on the large datasets DoclayNet and PublayNet,  we finally report the suboptimal configuration ours-Small instead.

\section{Conclusion and Future Work}
In this paper,  we propose a language-independent GNN framework for document layout analysis tasks. Our proposed model, Paragraph2Graph, uses a pre-trained CNN to encode image features and incorporates 2d OCR text coordinates and image features as node features in a graph. We use a dynamic graph convolutional neural network (DGCNN) to update the graph based on these features and include edge features based on relationships. To improve efficiency, we also propose a graph sampling method based on layout modality. Our method does not require a tokenizer or language model and can easily generalize to multilingual documents without modifications. We show that our method can achieve competitive results on three public datasets with fewer parameters. 
There are several potential improvements and attempts we leave for future work: (1) We have only experimented with only a few common GNNs,  while torch-geometric \cite{pytorch_geometric} officially offers nearly 60 related algorithms. Some of them can be a better substitution for DGCNN. (2) The backbone of image features can be pre-trained on document data making it better at capturing document image features. (3) Similar to the layoutLM which has a reasonable pre-training task to improve the merge of different modalities,  our model can be pre-trained with image reconstruction tasks as well, such as MAE. (4) Our model doesn't behave well on grouping tables and figures. Future research is needed to expand its generality on these important document layout components.

\section*{Acknowledgments}
We would like to acknowledge Xinxing Pan, Weihao Li, Binbin Yang, Hailong Zhang for their helpful suggestions.

%Bibliography
\bibliographystyle{unsrt}  
\bibliography{references}  

\section{Appendix}
\subsection{Graph Sampling Comparison}

\begin{figure}[!htbp]
\centering  %图片全局居中
\subfigure[ $\beta$-skeleton;\textbf{left}: $\beta$=1.0;\textbf{middle}: $\beta$=0.9;\textbf{right}: $\beta$=0.8. Decreasing $\beta$ can trade sparsity with connectivity, but a dense graph will make information propagation less effective. ]{
\label{Fig.1}
\includegraphics[width=0.3\linewidth]{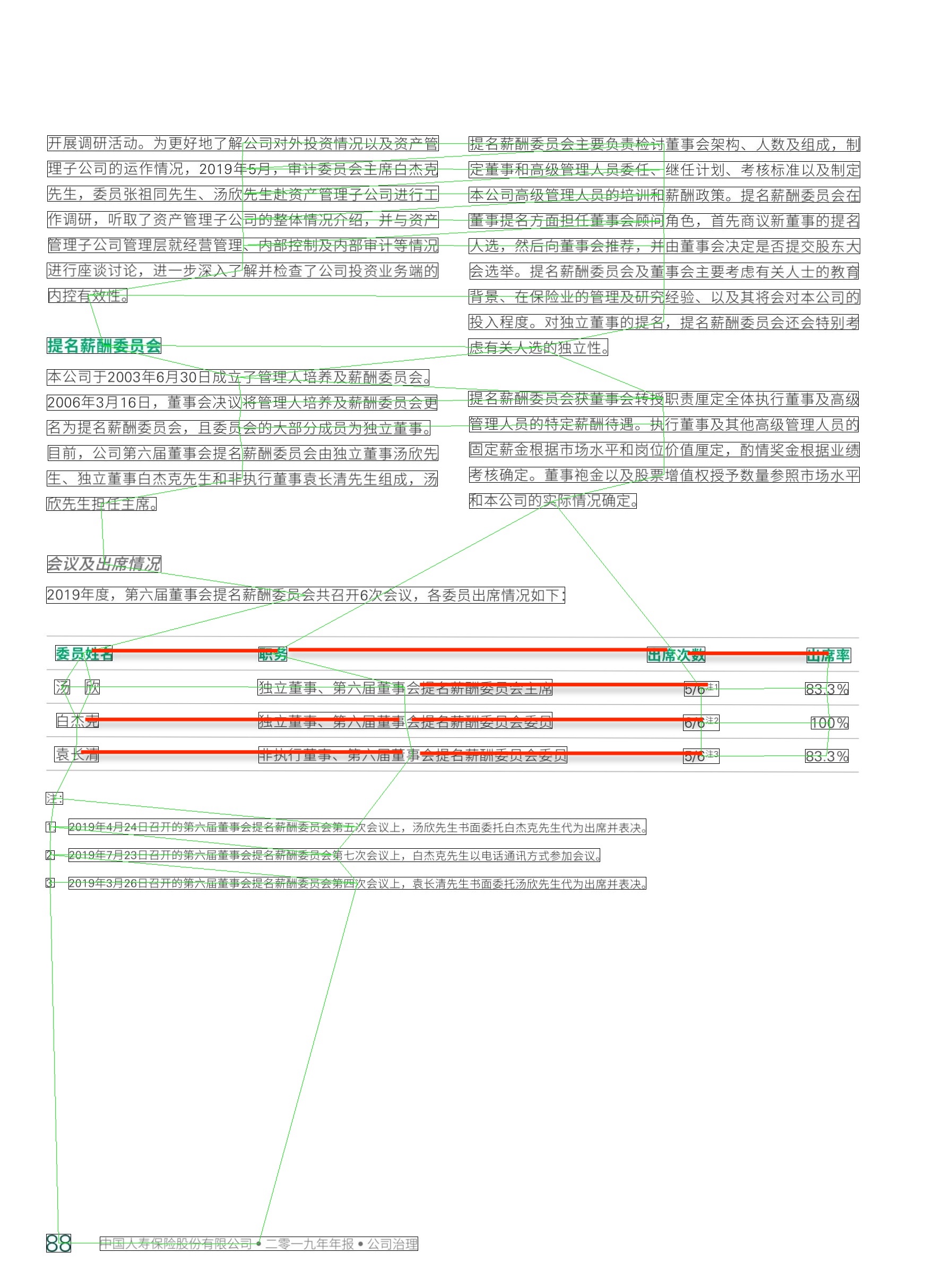}
\includegraphics[width=0.3\linewidth]{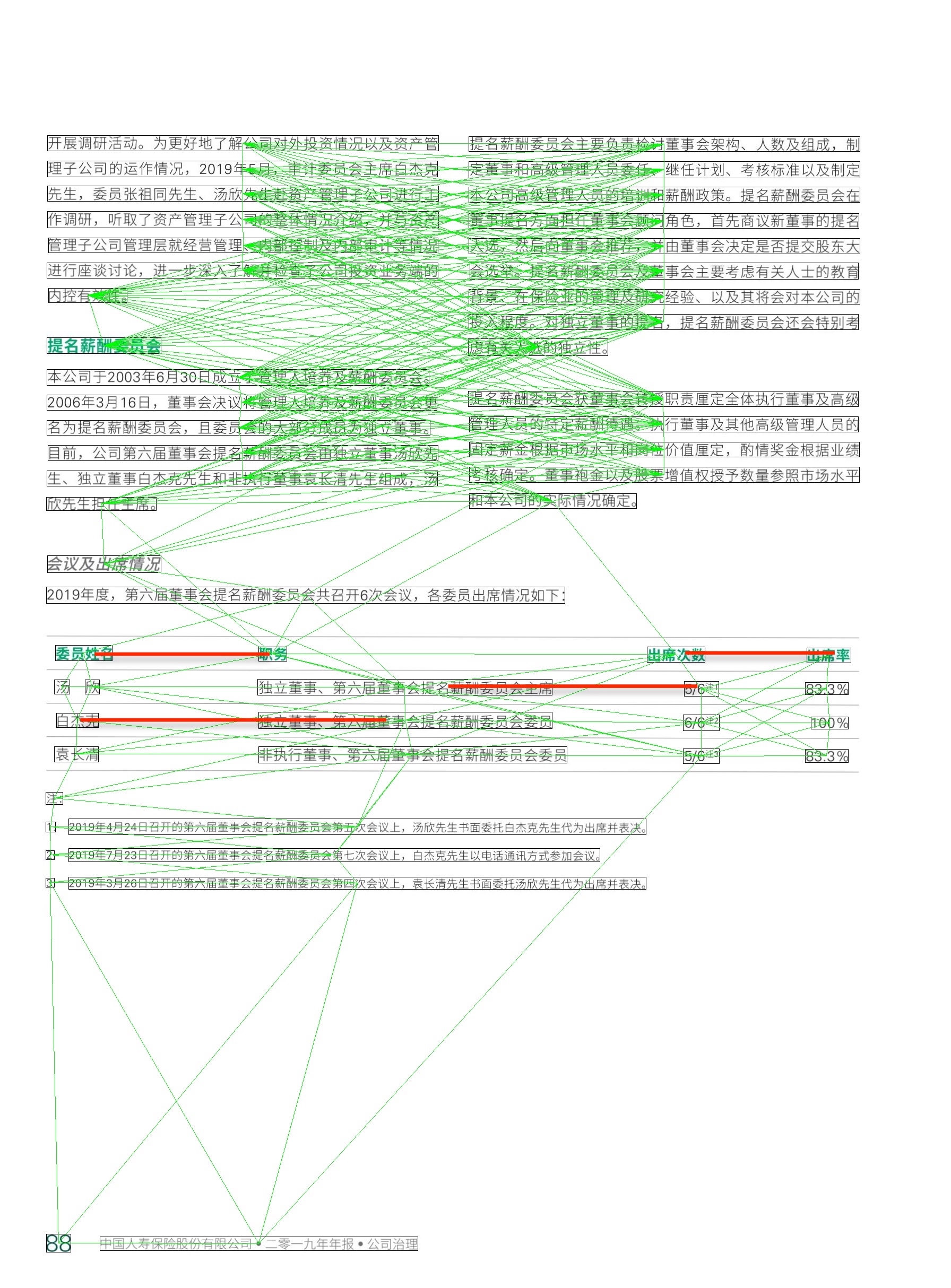}
\includegraphics[width=0.3\linewidth]{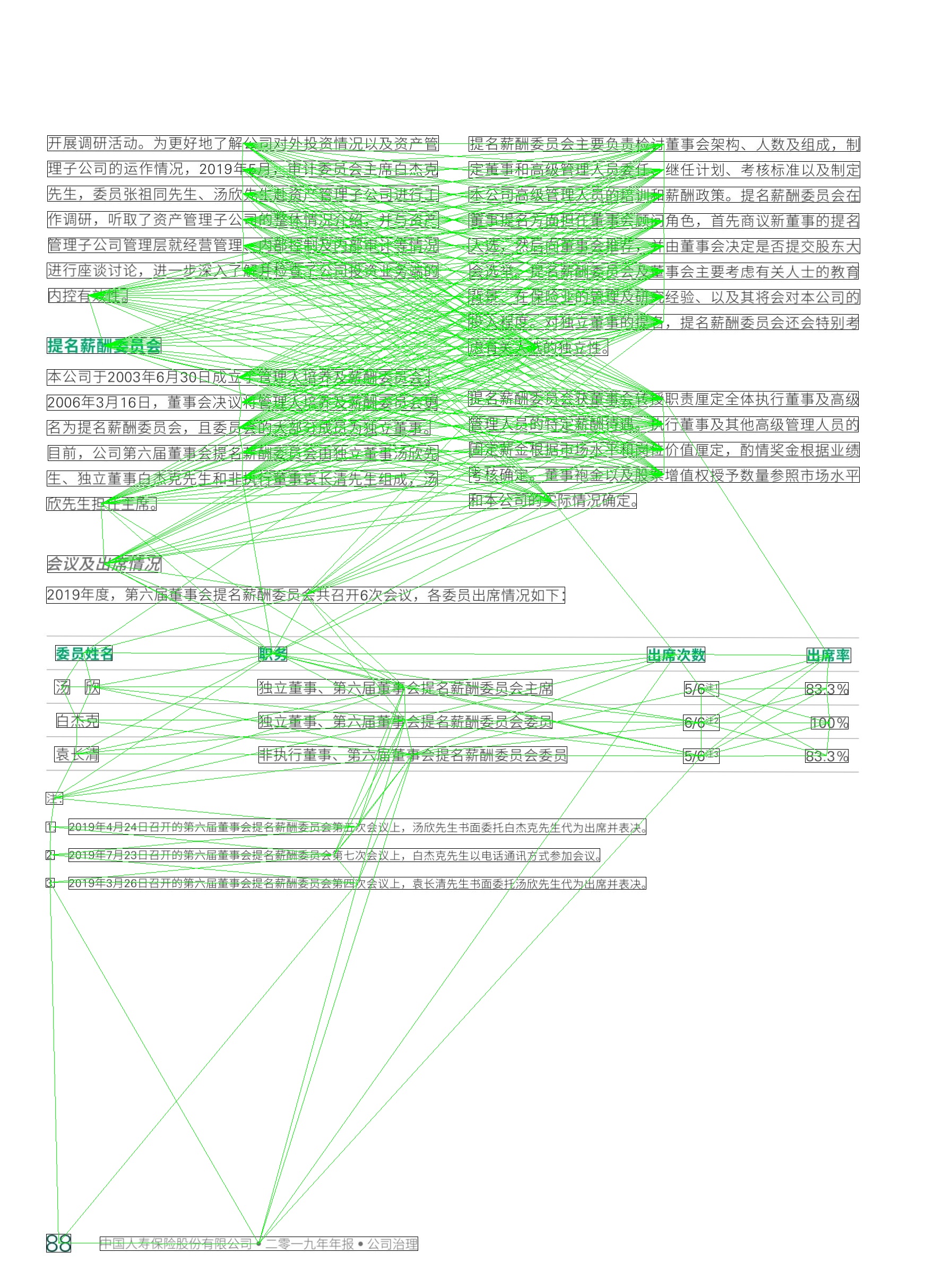}
}
\subfigure[KNN;\textbf{left}: $knn$=5;\textbf{middle}: $knn$=10;\textbf{right}: $knn$=20;Increasing $knn$ can mitigate connectivity missing, but make sampled graph much denser.]{
\label{Fig.2}
\includegraphics[width=0.3\linewidth]{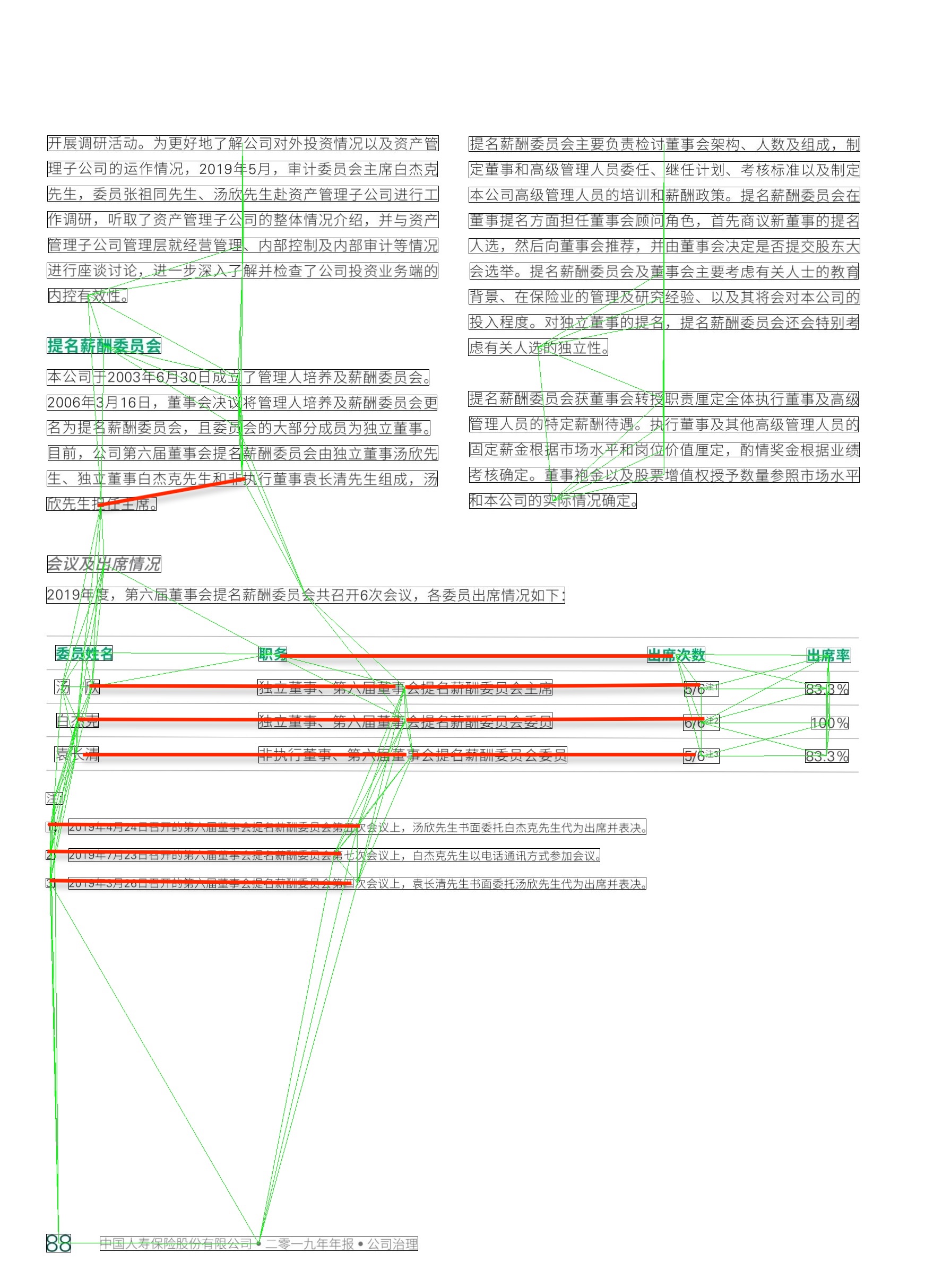}
\includegraphics[width=0.3\linewidth]{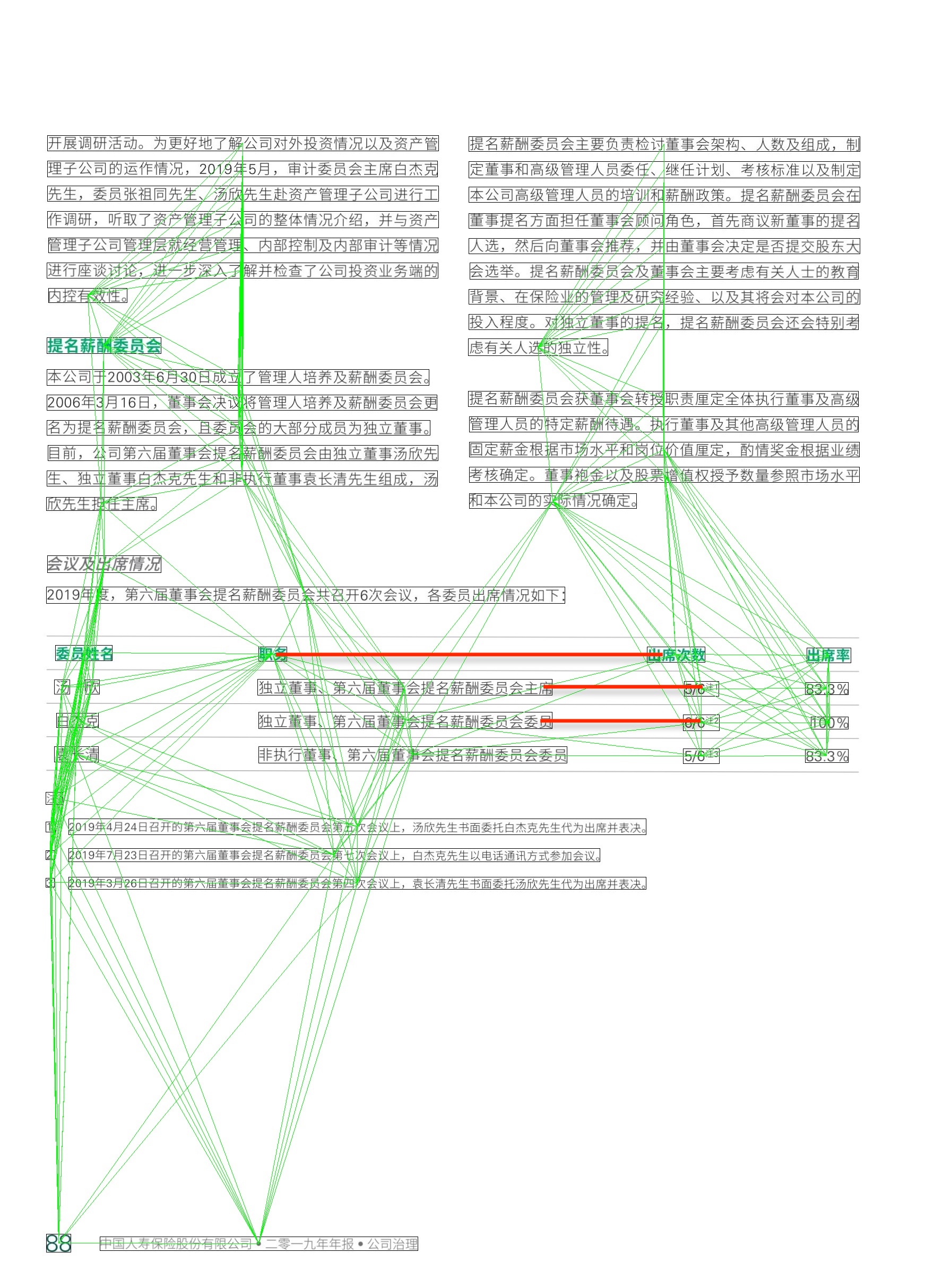}
\includegraphics[width=0.3\linewidth]{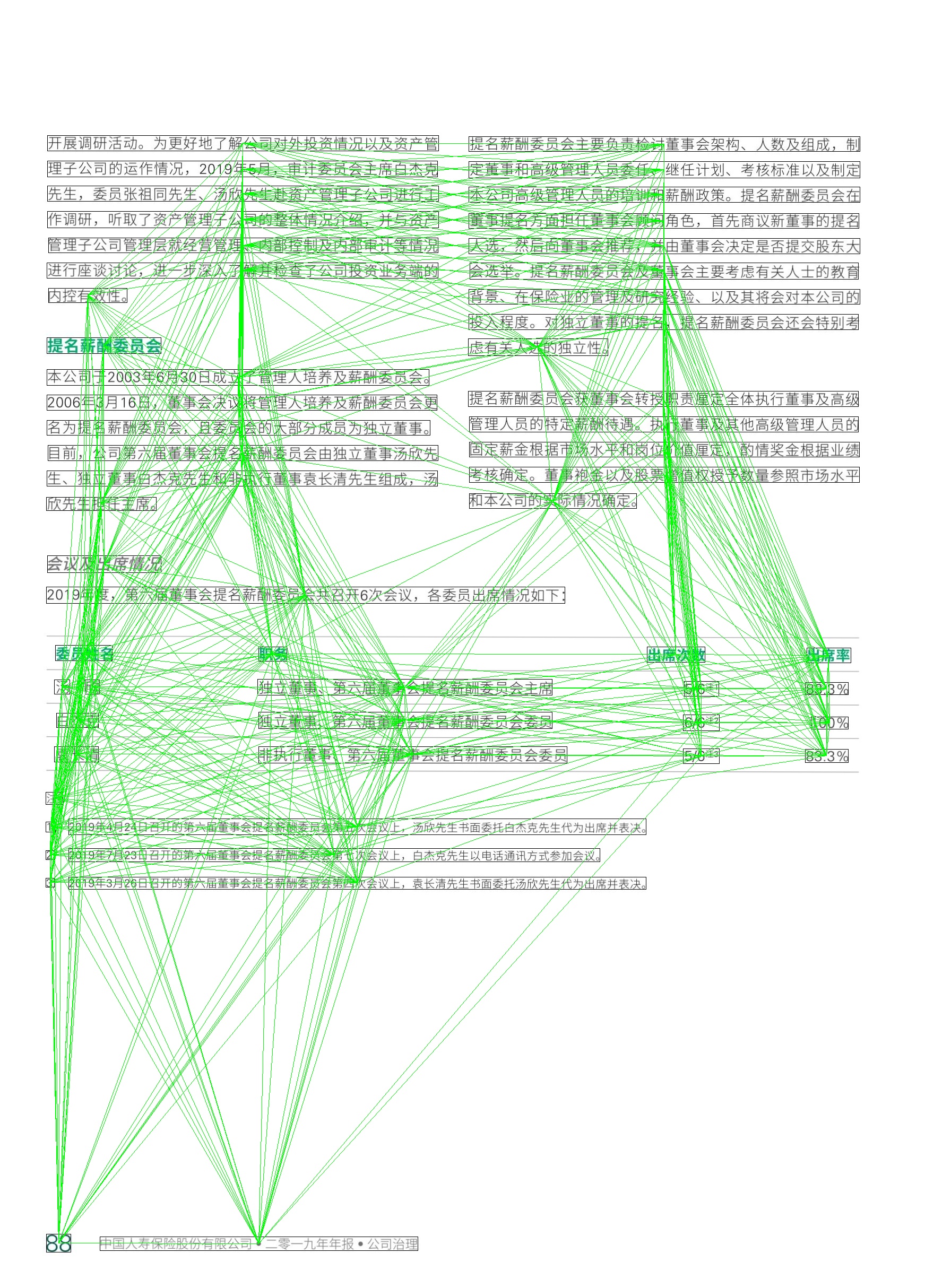}
}
\subfigure[ours]{
\label{Fig.3}
\includegraphics[width=0.3\linewidth]{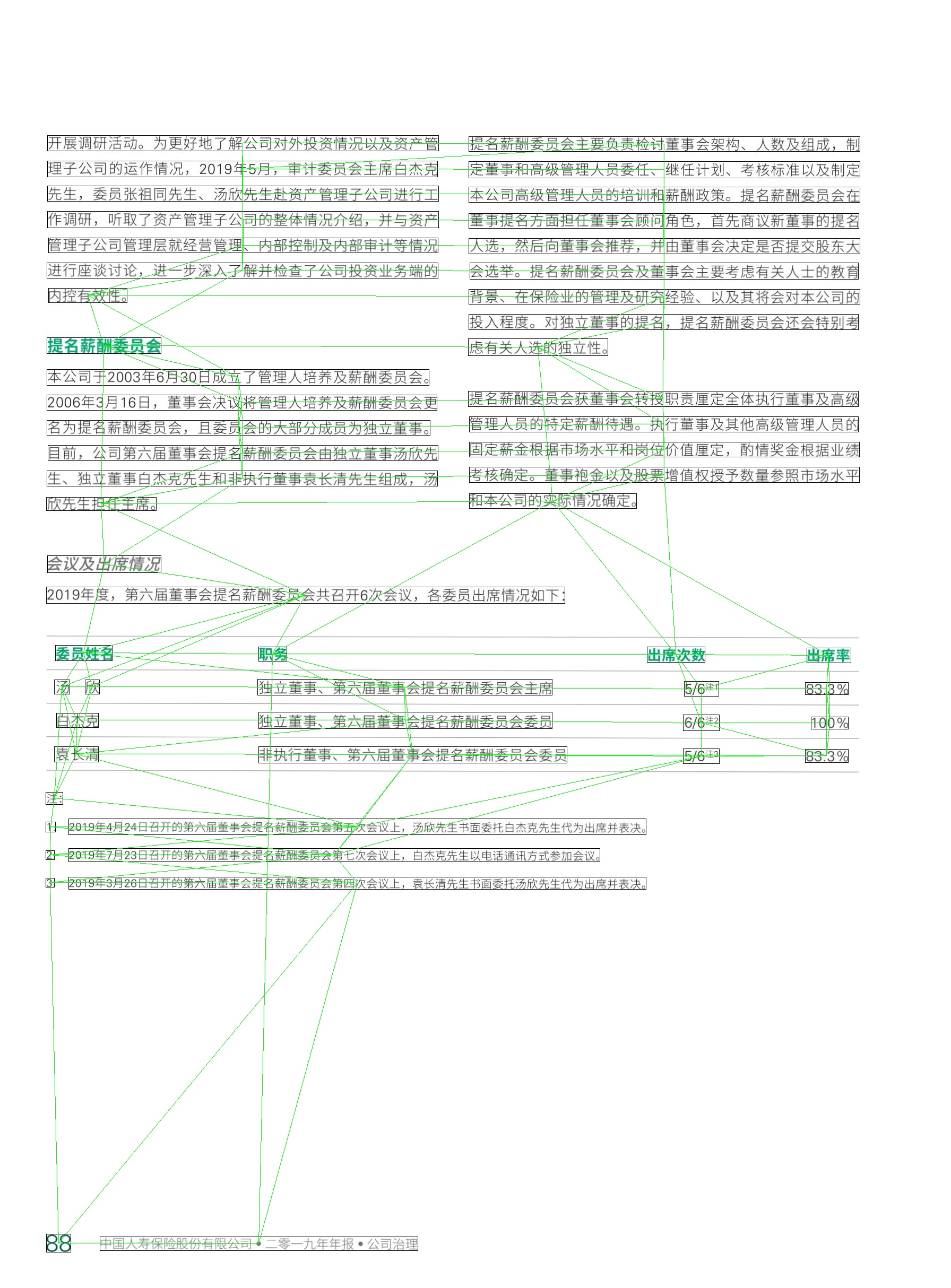}}
\caption{graph sampling results of KNN, $\beta$-skeleton and our strategy}
\label{fig:bad_case}
\end{figure}
As shown in Fig \ref{fig:bad_case}, green lines represent valid pairs connected by algorithms, red represents missing pairs that should be connected.The sampled results of KNN and $\beta$-skeleton are very sensitive to different parameters. For (a): when $\beta$=1, sampled graph miss many pairs, $\beta$=0.8 return enough pairs, but more negative pairs are introduced. Our sampling strategy can reach a balance between sparsity and connectivity.

\begin{table}[!htbp]
\centering
\begin{tabular}{|c|ccccc|c|c|c|}
\hline
mAP                           & Text           & Title          & List           & Table          & Figure         & Total           & Size \\\hline
Faster-RCNN\cite{Zhong2019PubLayNetLD}                   & 0.910          & 0.826          & 0.883          & 0.954          & 0.937          & 0.902                 & -      \\
Mask-RCNN\cite{Zhong2019PubLayNetLD}                     & 0.916          & 0.84           & 0.886          & 0.96           & 0.949          & 0.91                   & 168M   \\
Faster-RCNN{[}UDoc{]}\cite{Gu2021UniDocUP}         & 0.939          & 0.885          & 0.937          & 0.973          & 0.964          & 0.939                   & -      \\
Mask-RCNN{[}DiT-base{]}\cite{Li2022DiTSP}       & 0.934          & 0.871          & 0.929          & 0.973          & 0.967          & 0.935                    & 432M   \\
Cascade-RCNN{[}DiT-base{]}\cite{Li2022DiTSP}    & 0.944          & 0.889          & 0.948          & 0.976          & 0.969          & 0.945                     & 538M   \\
Cascade-RCNN{[}layoutlm-v3{]}\cite{huang2022layoutlmv3} & 0.945          & 0.906          & \textbf{0.955} & \textbf{0.979} & \textbf{0.970} & 0.951                   & 538M   \\
Faster-RCNN-Q\cite{Wang2021PostOCRPR}                 & 0.914          &                &                &                &                &                                & -      \\
Post-OCR\cite{Wang2021PostOCRPR}                      & 0.892          &                &                &                &                &                                & -      \\ \hline
ours-Small                          & \textbf{0.954}          & \textbf{0.913}          & 0.805          & 0.932          & 0.777          & 0.876                  & \textbf{77M}   \\\hline
\end{tabular}
\caption{performance on Publaynet on all categories.}
\label{table:Publaynet-all}
\end{table}

\begin{table}[!htbp]
\centering
\setlength{\parindent}{-4em}
\scalebox{0.7}{
\begin{tabular}{|c|c|c|c|c|c|}
\hline
mAP & Mask-RCNN-res50\cite{Pfitzmann2022DocLayNetAL}  & Mask-RCNN-resnext101\cite{Pfitzmann2022DocLayNetAL} & Faster-RCNN-resnext101\cite{Pfitzmann2022DocLayNetAL} & YOLO-v5x6\cite{Pfitzmann2022DocLayNetAL}   & ours-Small                       \\ \hline
Page-Header     & 0.719                                                           & 0.7                                                                  & 0.720                                                                  & 0.679                                                     & \textbf{0.796}  \\ 
Caption         & 0.684                                                           & 0.715                                                                & 0.701                                                                  & 0.777                                                     & \textbf{0.809}  \\ 
Formula         & 0.601                                                           & 0.634                                                                & 0.635                                                                  & 0.662                                                     & \textbf{0.726}  \\ 
Page-Footer     & 0.616                                                           & 0.593                                                                & 0.589                                                                  & 0.611                                                     & \textbf{0.920}  \\ 
Section-Header  & 0.676                                                           & 0.693                                                                & 0.684                                                                  & 0.746                                                     & \textbf{0.824}  \\ 
Footnote        & 0.709                                                           & 0.718                                                                & 0.737                                                                  & \textbf{0.772}                           & 0.625                            \\ 
Title           & 0.767                                                           & 0.804                                                                & 0.799                                                                  & \textbf{0.827}                           & 0.643                            \\ 
Text            & 0.846                                                           & 0.858                                                                & 0.854                                                                  & \textbf{0.881}                           & 0.827                            \\ 
List-Item       & 0.812                                                           & 0.808                                                                & 0.810                                                                  & \textbf{0.862}                           & 0.805                            \\ 
Picture         & 0.717                                                           & 0.727                                                                & 0.720                                                                  & \textbf{0.771}                           & 0.581                            \\ 
Table           & 0.822                                                           & 0.829                                                                & 0.822                                                                  & \textbf{0.863}                           & 0.559                            \\ \hline
Total-paragraph & 0.702                                                           & 0.7143                                                               & 0.7148                                                                 & 0.744                                                     & \textbf{0.771}  \\ \hline

Total           & 0.724                                                           & 0.735                                                                & 0.734                                                                  & \textbf{0.768}                           & 0.738                            \\ \hline
Params          & -                                                               & -                                                                    & 60M                                                                    & 140.7M                                                    & \textbf{19.95M} \\ \hline
\end{tabular}
}
\caption{performance on Doclaynet on all categories.}
\label{table:Doclaynet-all}
\end{table}

\end{document}